\documentclass{article}

\usepackage{microtype}
\usepackage{graphicx}
\usepackage{subfigure}
\usepackage{booktabs} 

\usepackage{hyperref}

\usepackage[accepted]{icml2024}

\usepackage{amsmath}
\usepackage{amssymb}
\usepackage{mathtools}
\usepackage{amsthm}

\usepackage[capitalize,noabbrev]{cleveref}

\theoremstyle{plain}

\theoremstyle{definition}

\theoremstyle{remark}

\usepackage[textsize=tiny]{todonotes}

\icmltitlerunning{Evaluating Large Language Models for Generalization and Robustness via Data Compression}

\begin{document}

\twocolumn[
\icmltitle{Evaluating Large Language Models for \\ Generalization and Robustness via Data Compression}



\icmlsetsymbol{equal}{*}

\begin{icmlauthorlist}
\icmlauthor{Yucheng Li}{surrey}
\icmlauthor{Yunhao Guo}{HEU}
\icmlauthor{Frank Guerin}{surrey}
\icmlauthor{Chenghua Lin}{manchester}
\end{icmlauthorlist}

\icmlaffiliation{surrey}{University of Surrey, UK}
\icmlaffiliation{manchester}{University of Manchester, UK}
\icmlaffiliation{HEU}{Harbin Engineering University, China}

\icmlcorrespondingauthor{Yucheng Li}{yucheng.li@surrey.ac.uk}

\icmlkeywords{Machine Learning, ICML}

\vskip 0.3in
]




\printAffiliations{}

\begin{abstract}

Existing methods for evaluating large language models  face challenges such as data contamination, sensitivity to prompts, and the high cost of benchmark creation. To address this, we propose a lossless data compression based evaluation approach that tests how models' predictive abilities generalize after their training cutoff.
Specifically, we collect comprehensive test data spanning 83 months from 2017 to 2023 and split the data into training and testing periods according to models' training data cutoff. We measure: 1) the compression performance on the testing period as a measure of generalization on unseen data; and 2) the performance gap between the training and testing period as a measure of robustness.
Our experiments test 14 representative large language models with various sizes on sources including Wikipedia, news articles, code, arXiv papers, and multi-modal data. We find that the compression rate of many models reduces significantly after their cutoff date, but models such as Mistral and Llama-2 demonstrate a good balance between performance and robustness. Results also suggest that models struggle to generalize on news and code data, but work especially well on arXiv papers. We also find the context size and tokenization implementation have a big impact of on the overall compression performance. Our data and code can be found at \url{https://github.com/liyucheng09/llm-compressive}.

\end{abstract}

\section{Introduction}

Modern evaluation methods for large language models typically use comprehensive benchmark suites that include a broad range of tasks and domains. This aims to assess the generalization ability of different models in varying scenarios and prevent overfitting on any specific task or dataset \cite{chang2023survey}. For example, the GLUE benchmark \cite{wang2018glue,wang2019superglue} consists of 10+ NLP tasks and is widely used in benchmarking large language models \cite{zhang2022opt,chowdhery2023palm}. More recently, GPT-4 was tested on 30+ human academic tests and 10+ NLP benchmarks \cite{achiam2023gpt}.

However, this methodology suffers from three critical problems. Firstly, benchmark results depend heavily on prompts \cite{sclar2023quantifying}. Different prompts for the same model can produce wildly different results. This makes it hard to evaluate the models directly, without the evaluation being inflated by clever prompt engineering. For example, Google Gemini outperforms baseline models with customized prompts, but fails to beat other models with simple few-shot prompts \cite{team2023gemini}, making it difficult to ascertain whether the good performance is the result of model abilities or well-designed prompts.
Secondly, the increasing scale of pre-training data raises the possibility of including benchmark data in the pre-training stage of large language models, leading to the \textit{data contamination} issue \cite{jacovi2023stop,sainz2023nlp}. Data contamination allows models to achieve higher metrics through ``memorization'', rather than demonstrating true generalization. This undermines the reliability of model comparisons and can also mislead model development. Studies estimate that 30-80\% of examples in popular benchmarks like MMLU \cite{hendrycks2020measuring} and SQuAD \cite{rajpurkar2018know} are contaminated, with the ratio increasing rapidly over time \cite{li2023open}.
Finally, due to the human efforts required, constructing new and diverse benchmarks is very expensive. This prevents regular updates of existing benchmarks and further diversification to cover more domains and tasks. For example, benchmarks like MMLU and C-Eval \cite{huang2023ceval} are collected from human academic tests which have accumulated over the past 10 years to reach their current scale. Continuing with the same method of construction will make it extremely difficult to collect new and updated tests in the near future. As a result, the community urgently needs new methods for assessing large language models.

In this paper, we propose a novel approach that evaluates base large language models though data compression across a wide range of time periods. Firstly, we use lossless data compression as the metric to assess model performance. Compression metrics are widely used in benchmarking language modeling \cite{radford2019language,dai2019transformerxl} and are shown to be strongly correlated with generalization ability and in-context learning performance \cite{deletang2023language,rae_2023_compression}. As compression requires models to work on raw data, our method avoids the potential interference introduced by prompts and also reduces human effort in creating benchmark questions. Secondly, similar to the idea of the train/test data split in traditional machine learning, we split our testing data into the \textit{training period} and the \textit{testing period} according to models' cutoff dates. Models are first evaluated on data from the training period, which is regarded as the in-distribution performance. They are then evaluated on data from the testing period (i.e., data emerging after the training concludes). This focuses on the models' generalization to new, unseen data, and aligns well with how language models need to handle new data in real world scenarios. Finally, we use the gap between the training and testing periods as a measure of model robustness over time, similar to how we compare training error and testing error in traditional machine learning. A small difference between the training and testing periods indicates a strong robustness of model performance, while a large gap may signal overfitting.

We collect test data spanning 83 months, from Jan 2017 to Nov 2023, to analyze how models generalize over time. This data covers a broad range of domains and also multi-modal sources known for their time-sensitivity, including Wikipedia pages, GitHub code, BBC news, arXiv papers, news images, and audio.
Our experiments cover a diverse range of foundation models including LLaMA \cite{touvron2023llama}, Llama-2 \cite{touvron2023llama2}, CodeLlama \cite{roziere2023codellama}, Yi \cite{Yi}, Mistral \cite{jiang2023mistral}, Baichuan2 \cite{yang2023baichuan}, InternLM \cite{team2023internlm}, ChatGLM \cite{du2021glm}, and Qwen \cite{bai2023qwen} across multiple model sizes (6B, 7B, 13B, 34B, 65B, and 70B parameters).
We also discuss the impact of various model context sizes (2K, 4K, and 8K) and different tokenization approaches. Finally, we compare results of our method to those of established benchmarks such as HumanEval \cite{chen2021evaluating} and MMLU.

Our key findings are as follows: 1) Models' compression performance over time correlates closely with their training data collection time, with clear divergences after the cutoff date. 2) Models with similar performance on the training period can demonstrate widely different generalization results on new, unseen data. 3) Generalization difficulty varies across test datasets. Models struggle to generalize on wikitext, news, and code, but generalize well on arXiv content. 4) All models fail to compress multi-modal data, showing limited capabilities in dealing with raw byte streams. 5)~Larger contexts generally lead to better performance but do not exceed a small context + sliding window approach. 6)~Models with larger vocabularies in tokenization have more difficulty with token-level prediction.

\section{Background}

\subsection{Language Models Evaluation}

Traditionally, language model performance is measured by intrinsic metrics such as negative log-likelihood (NLL) loss, perplexity, and bits-per-character (BPC). These metrics primarily focus on a model's ability to predict or generate text sequences. For example, NLL loss is defined as follows.
\begin{equation}
L(x_{1:n}) = -\log P(x_{1:n}) = - \sum_{t=1}^n \log P(x_t|x_{1:t-1})
\label{nll}
\end{equation}
where the likelihood \(P(x_{1:n})\) is expanded to \(P(x_t|x_{1:t-1})\) based on the chain rule. It quantifies how well the model predicts a given text sequence, with lower NLL loss indicating better predictive accuracy. These language modeling metrics are typically used together with established datasets. For example, GPT-2 \cite{radford2019language} was evaluated on Wikitext-2 \cite{merity2016pointer}, Penn Tree Bank \cite{marcus1994penn}, and enwik8 \cite{Hutter2006prize} using perplexity and BPC as metrics. More recently, as language models are increasingly being used in NLP tasks and real-world applications, they are often evaluated based on how well they perform on downstream tasks 
\cite{beeching2023open,2023opencompass}. One common way to perform task-based evaluation is to further fine-tune and test models on task-specific datasets, which is usually used for BERT \cite{devlin2018bert} and RoBERTa \cite{liu2019roberta} models. In contrast, another emerging way is to use pure prompting to get models' responses. This method is more frequently used in modern large language models such as GPT-4, PaLM and LLaMA. In this case, the evaluation also takes the models' in-context learning ability into consideration, which refers to how well the models can learn from the information and demonstrations provided in the limited context.

However, as language models grow in size and consume more data, both methods face challenges. Firstly, as both methods rely on internet-sourced test datasets, concerns of data contamination arise. This refers to evaluating on examples that are explicitly or implicitly included in the training data. For example, GPT-3 reported that Wikipedia-based language modeling benchmarks and the Children's Book Test dataset are almost entirely contained in its training data, with other Wikipedia-based reading comprehension benchmarks such as QuAC, SQuADv2 and DROP showing over 90\% contamination \cite{brown2020language}. \citet{roberts2023data} conduct a longitudinal analysis of the programming ability of large language models, revealing a significant association between a code problem's presence on GitHub and a model's pass rate for that problem. \citet{li2023open} also show that benchmark contamination can inflate accuracy by 7\% to 14\% on benchmarks like MMLU and C-Eval, even when contamination only includes questions without revealing the associated correct answers. 
Secondly, large language models have been shown to be highly sensitive to prompt design \cite{sclar2023quantifying}.  There are also prompting techniques such as Chain-of-Thought \cite{wei2022chain}, and Lottery prompting \cite{chen2023Lottery} which use this sensitivity and are designed to enhance model performance on various tasks.
This complicates the accurate measurement of model performance when relying solely on the pure prompting approach.

\subsection{Compression and Language Models}
It is well established that compression is essentially prediction, which effectively links compression and langauge models \cite{deletang2023language}. The source coding theory from Shannon's information theory \cite{shannon1948mathematical} suggests that the number of bits required by an optimal entropy encoder to compress a message \(x_{1:n}\) is equal to the negative \(\log_2\) likelihood of the message given by a statistical model (i.e., \(-\log_2 P(x_{1:n})\)). Comparing this to Eq.~\ref{nll}, we find that optimizing the NLL loss of language modeling and optimizing the expected message length to compress \(x_{1:n}\) is fundamentally equivalent. Therefore, prior work demonstrates that language models and other neural networks can be used to achieve state-of-the-art lossless compression via arithmetic coding \cite{yang2023introduction,valmeekam2023llmzip}. In addition, \citet{deletang2023language} and \citet{rae_2023_compression} further argue that the ability to compress information is closely related to the ability to generalize. Finally, \citet{sutskever2023generalization} leverage the ability of compression to explain the effectiveness of unsupervised learning.

\section{Our Method}

In this section, we explain how to perform data compression with a language model. There are typically two stages involved: 1) calculating the likelihood of the data \(X\) with the language model \(f\); and 2) applying an entropy encoding algorithm, usually arithmetic coding, which represents \(X\) based on the computed probabilities to reduce its overall size. Firstly, we apply language models to the data to be compressed \(X\), to obtain the probability distribution \(P(X)\). For textual data, consider we have a language model \(f\) and a textual dataset \(X = (x_0, \cdots, x_n)\) consisting of a stream of tokens. The likelihood is computed as follows:
\begin{equation}
    f(X) = P(X) = \prod_{i=1}^n P(x_i|x_{1:i-1})
\end{equation}
However, due to modern Transformer-based language models having a limited context window denoted as \(C\), we need to segment the dataset \(X\) into chunks of size \(C\) and feed them to the language models one-by-one. As a result, the actual likelihood is computed as follows:
\begin{equation}
    f(X) = P(X) = \prod^N\prod_{j=1}^C P(x_j|x_{1:j-1})
    \label{likelihood}
\end{equation}
where \(N\) indicates the number of chunks. It is rather straightforward to further apply language models to multi-modal data. In this case, \(X\) consists a stream of bytes instead of tokens. Each byte in \(X\), ranging from \texttt{<0x00>} to \texttt{<0xFF>}, is individually mapped to a unique input ID that the tokenizer reserves for byte characters. The likelihood of multi-modal \(X\) is then calculated based on Eq.~\ref{likelihood}, the  same as for text data.

\begin{algorithm}[t]
\caption{Arithmetic Coding}
\label{Arithmetic Coding}
\begin{algorithmic}
\REQUIRE A stream of bytes $x_{1:n}$ and a language model $f$
\ENSURE Encoded byte stream $E$

\STATE $low \gets 0.0$
\STATE $high \gets 1.0$
\FOR{$x_i$ in $x_{1:n}$}
    \STATE $range \gets high - low$
    \STATE $high \gets low + range \times P(x_{1:i})$
    \STATE $low \gets low + range \times P(x_{1:i-1})$
\ENDFOR
\STATE $E \gets$ value in $[low, high)$
\STATE \textbf{return} $E$
\end{algorithmic}
\end{algorithm}

\begin{algorithm}[t]
\caption{Arithmetic Decoding}
\label{Arithmetic Decoding}
\begin{algorithmic}
\REQUIRE Encoded byte stream $E$, length of $x_{1:n}$, and the same language model $f$
\ENSURE Decoded byte stream $x_{1:n}$

\STATE $low \gets 0.0$
\STATE $high \gets 1.0$
\STATE $x \gets$ empty byte stream
\FOR{$i = 1$ to $n$}
    \STATE $range \gets high - low$
    \FOR{each possible byte $x$}
        \STATE $high \gets low + range \times P(x_{1:i})$
        \STATE $low \gets low + range \times P(x_{1:i-1})$
        \IF{$code$ is in the range of $[low, high)$}
            \STATE append $x$ to $x_{1:i-1}$
            
            \STATE \textbf{break}
        \ENDIF
    \ENDFOR
\ENDFOR
\STATE \textbf{return} $D$
\end{algorithmic}
\end{algorithm}

Secondly, dataset \(X\) and its likelihood \(P(X)\) are given to an entropy encoding algorithm to obtain the final compressed binary representation. We use arithmetic coding in our experiments, which is known to be optimal in terms of coding length.
The encoding and decoding process of arithmetic coding with a language model are shown in Algorithms \ref{Arithmetic Coding} and \ref{Arithmetic Decoding} \cite{rissanen1976generalized,pasco1976source,Nelson14DataCompression}. As the language model is working on chunks, the arithmetic coding is also performed on chunked batches and then the compressed data is concatenated into the final result. Assuming the arithmetic operations were implemented with infinite precision, the arithmetic encoding can reach the length \(-\log_2\lceil P(X)\rceil + 1\), which is  close to the theoretical optimal length \(-\log_2 P(X)\) \cite{deletang2023language}. A practical implementation will introduce inefficiency of \(O(n2^{-B})\) bits to the final arithmetic code, where \(B\) indicates the precision of the implemented arithmetic operations \cite{howard1992analysis}. Our implementation is based on 32-bit arithmetic, in which case the overhead is negligible.

\section{Experiment}

\begin{table}
    \centering
    \resizebox{\columnwidth}{!}{
    \begin{tabular}{lcccc}
\toprule
        Dataset & Modality & \#Docs & Size & Total size \\
        \midrule
        Wikitext & Text & 500 & 23MB & 1.9GB \\
        BBC News & Text &  1,270 & 6MB & 490MB \\
        GitHub & Text & 395 & 35MB & 2.8GB \\
        ArXiv & Text & 533 & 28MB & 2.3GB \\
        BBC Images & Bytes & 1,589 & 12MB & 1GB \\
        Audio-Mix & Bytes & 103 & 12MB & 1GB \\
    \bottomrule
    \end{tabular}}
    \caption{Test data to compress in our experiments. \#Docs and Size are averaged per month.}
    \label{tab:data}
    \vspace{-0.8em}
\end{table}

\subsection{Data Collection}

We use a mixture of text, image, and audio sources in our experiments. As discussed in \citet{deletang2023language}, language models, when applied on multi-modal data, are asked to identify useful patterns from the limited context for compression. Thus, evaluating compression capability on images and audio can serve as an effective measure of the models' in-context learning and generalization abilities. We aim to cover a diverse set of domains that are known for their time-sensitivity. Detailed statistics are reported in Table \ref{tab:data}. Our test data is collected spanning 83 months, from Jan 2017 to Nov 2023. All sources here are under an open source license or are freely accessible for research purposes (see Appendix \ref{license}).

\begin{table}[t]
    \centering
    \resizebox{\columnwidth}{!}{
    \begin{tabular}{lcccc}
    \toprule
        Model & Release & Cutoff & Size & Context \\
        \midrule
        LLaMA & 2023-02 & 2020\textsuperscript{*}  &  7/13/65B & 2048 \\
        InternLM & 2023-06 & - & 7B & 2048 \\
        Llama-2 & 2023-07 & 2022\textsuperscript{†} & 7/13/70B & 4096 \\
        CodeLlama & 2023-08 & - & 7B & 16K \\
        Baichuan2 & 2023-09& - & 7B & 4096\\
        Mistral & 2023-09&- & 7B & 32K \\
        Qwen & 2023-09&- & 7B & 32K\\
        ChatGLM3 & 2023-10 &- & 6B & 32K\\
        Yi & 2023-11&- & 7/34B & 4K/200K\\
        \bottomrule
    \end{tabular}}
    \caption{Models included in our experiments. Cutoff indicates their knowledge cutoff date. \textsuperscript{*}\citet{touvron2023llama} do not include an exact knowledge cutoff date, but most of their pre-training data uses CommonCrawl dumps from 2017-2020. \textsuperscript{†}The model card of Llama-2 states that the pre-training cutoff for the base model is September 2022 \cite{touvron2023llama2}.}
    \label{tab:models}
    \vspace{-0.8em}
\end{table}

\noindent\textbf{English Wikipedia.}
English Wikipedia (also referred to as Wikitext below) is a high-quality material for testing encyclopedic knowledge. Unlike previous studies that collect data from a specific Wikipedia dump, we monitor 500 specific articles and document their version monthly. This provides us with an evolving timeline allowing us to analyze how large language models generalize over time. The average length of these articles is 55K characters per article in 2017, which is increased to 61K in 2023.

\noindent\textbf{BBC News Articles.}
News is a well-known time sensitive information source. When compared to encyclopedic data, the topic and content of news articles can change more randomly and dramatically (see Appendix \ref{how different}). This source aims to assess large language models' generalization on such a rapidly changing domain. We only collect articles that have appeared on the front page of BBC to ensure that they are representative. The average length of news articles is 12K characters per article.

\noindent\textbf{Github Code.}
This source aims to assess models' generalization on programming code. We monitor 75 popular GitHub projects that have rich commit history. The test data only includes newly added files and files that have changed dramatically (more than 50\%) in each month.

\begin{figure*}[t]
    \centering
    \includegraphics[width=\textwidth]{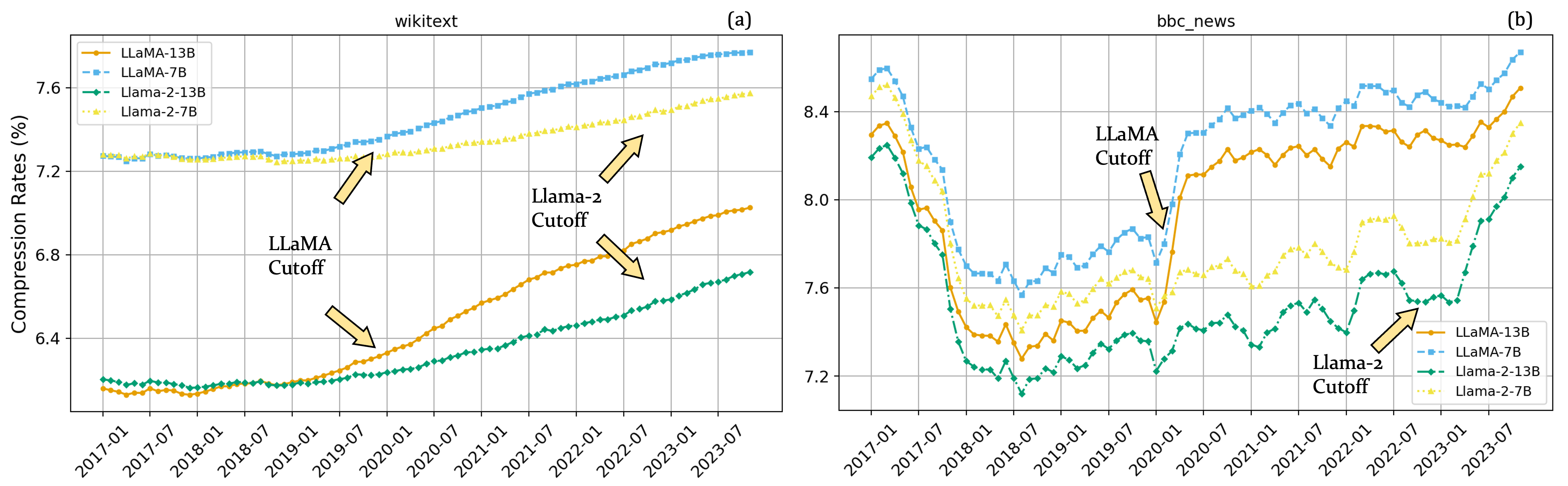}
    \vspace{-0.8em}
    \caption{The correlation between model compression rate (\%, lower is better) and their cutoff date. The cutoff for LLaMA and Llama-2  is 2020 (estimated) and September 2022, respectively (see \S\ref{models_and_metrics} for details).}
    \label{fig:llama-time}
\end{figure*}

\noindent\textbf{ArXiv.}
We use arXiv papers to test models' generalization on scientific data. We randomly collect papers from all disciplines. Author information, bibliographies, and appendices are excluded from the LaTeX source file and only main content starting from the introduction is used in our experiments.

\noindent\textbf{BBC Images.}
Images in BBC news articles are used in our testing. We extract contiguous patches of size 64 × 128 from all images. Then, these patches are flattened and converted to grayscale so that each byte represents exactly one pixel. When dealing with byte streams in large language models, we simply use the reserved \textit{ token\_ids} for bytes in the tokenizer. Each byte is mapped to a unique \textit{input\_id} and the output space is also adjusted accordingly to the byte space of size 256. We collect 12MB of data for each month, and ensure the total size is 1GB.

\noindent\textbf{Audio-Mix}
Audio from BBC radio and podcasts are used as our audio source. All audio files are converted to 16kHz FLAC format from neither m4p or mp3 format. Similar to images, audio samples are chunked into byte batches. We randomly sample 12MB of data per month, reaching a total size of 1GB.

\subsection{Models and Metrics}
\label{models_and_metrics}

We evaluate a wide range of open-source foundation models in our experiments, with details reported in Table \ref{tab:models}. As we are assessing models' generalization over time, we also include their release date and knowledge cutoff date. Although many of these models have their technical reports released, limited details are shared regarding their pre-training data. \citet{touvron2023llama} reported that the pre-training data for the LLaMA models primarily consists of CommonCrawl dumps from 2017 to 2020, which contribute 82\% of the data. The remaining 18\% is composed of content from GitHub, arXiv, and Wikipedia. Llama-2 models extend the pre-training data of LLaMA with more up-to-date information, but we are not aware of the precise composition. The exact knowledge cutoff date is September 2022 \cite{touvron2023llama2}.
All models listed are base language models without supervised fine-tuning or reinforcement learning. All results presented in the next section are obtained with a 2K context window unless otherwise stated. The selected models are predominantly general-purpose language models, except for CodeLlama, which is a code-specialized foundation model developed by further training Llama-2 on code-specific datasets.
In addition to language models, we also involve Gzip \cite{deutsch1996gzip}, PNG \cite{png}, and FLAC \cite{flac} to provide a comparative perspective against traditional compression methods. Compression rate (compressed size / raw size) is used as the primary metric to assess the compression performance of large language models.

\section{Results}

In this section, we first discuss how the performance of large language models correlates with their training cutoff date. We then evaluate large language models with respect to their generalization capability and robustness and present a model-wise comparison. We also compare our compression-based metric to established benchmarks. In addition, we analyze how models generalize differently on different testing datasets. Finally, we analyze how context size and tokenizer implementation affect the compression rate of large language models.

\begin{table*}[t]
    \centering
    \resizebox{\textwidth}{!}{
\begin{tabular}{lcccccccccccc}
\toprule
{} & \multicolumn{2}{c}{Wikitext} & \multicolumn{2}{c}{BBC News} & \multicolumn{2}{c}{Code} & \multicolumn{2}{c}{arXiv} & \multicolumn{2}{c}{Image} & \multicolumn{2}{c}{Audio} \\
\cmidrule(lr){2-3}
\cmidrule(lr){4-5}
\cmidrule(lr){6-7}
\cmidrule(lr){8-9}
\cmidrule(lr){10-11}
\cmidrule(lr){12-13}
Model &     Avg. &           2023 &     Avg. &           2023 &   Avg. &           2023 &   Avg. &           2023 &    Avg. &            2023 &    Avg. &            2023 \\
\midrule
Baichuan2-7B &    7.825 &  7.932 ↑ .124 &    8.092 &  8.385 ↑ .339 &  4.118 &  4.293 ↑ .202 &  9.118 &  9.018 ↓ .115 &  169.3 &  \underline{166.2 ↓ 3.58} &  177.3 &  177.2 ↓ .125 \\
CodeLlama-7B &    9.394 &  9.371 ↓ .027 &    9.371 &  9.565 ↑ .223 &  3.054 &  \textbf{3.479 ↑ .490} &  10.05 &  9.889 ↓ .184 &  162.1 &  \textbf{159.5 ↓ 2.95} &  146.3 &  \textbf{150.0 ↑ 4.26} \\
Chatglm3-6B  &    8.177 &  8.314 ↑ .158 &    8.401 &  8.726 ↑ .375 &  4.640 &  4.712 ↑ .082 &  9.638 &  9.500 ↓ .158 &  181.0 &  174.1 ↓ 8.01 &  183.5 &  187.9 ↑ 5.01 \\
Internlm-7B  &    9.969 &  9.992 ↑ .027 &    9.694 &  9.955 ↑ .301 &  4.623 &  4.924 ↑ .347 &  11.00 &  10.88 ↓ .136 &  180.0 &  178.9 ↓ 1.26 &  212.1 &  209.3 ↓ 3.26 \\
LLaMA-7B     &    7.463 & 7.752 ↑ .333 &    8.186 &  8.532 ↑ .398 &  4.194 &  4.448 ↑ .292 &  9.266 &  9.186 ↓ .091 &  194.7 &  190.9 ↓ 4.44 &  198.7 &  201.7 ↑ 3.41 \\
Llama-2-7B   &    7.349 &  \underline{7.539 ↑ .219} &    7.794 &  \underline{8.107 ↑ .361} &  4.339 &  4.455 ↑ .134 &  9.197 &  9.088 ↓ .126 &  182.3 &  179.5 ↓ 3.28 &  162.6 &  \underline{166.7 ↑ 4.66} \\
Mistral-7B   &    7.542 &  \textbf{7.642 ↑ .115} &    7.484 &  \textbf{7.827 ↑ .396} &  3.964 &  4.043 ↑ .091 &  8.452 &  \textbf{8.372 ↓ .092} &  202.4 &  198.8 ↓ 4.19 &  195.3 &  200.0 ↑ 5.37 \\
Qwen-7B      &    7.505 &  7.768 ↑ .304 &    8.158 &  8.472 ↑ .361 &  3.679 &  \underline{3.858 ↑ .206} &  9.006 &  \underline{8.911 ↓ .109} &  \multicolumn{2}{c}{-} &  \multicolumn{2}{c}{-} \\
Yi-6B        &    7.897 &  8.109 ↑ .244 &    7.969 &  8.264 ↑ .340 &  4.492 &  4.545 ↑ .061 &  9.144 &  9.043 ↓ .117 &  192.1 &  187.6 ↓ 5.17 &  197.0 &  201.0 ↑ 4.70 \\
\midrule
LLaMA-13B    &    6.488 &  6.979 ↑ .566 &    7.959 &  8.359 ↑ .461 &  4.013 &  4.296 ↑ .326 &  9.003 &  8.934 ↓ .080 &  176.0 &  \underline{171.2 ↓ 5.55} &  177.0 &  179.2 ↑ 2.21 \\
Llama-2-13B  &    6.342 &  6.658 ↑ .364 &    7.528 &  7.886 ↑ .413 &  4.138 &  4.288 ↑ .172 &  8.930 &  8.838 ↓ .106 &  158.5 &  \textbf{155.6 ↓ 3.37} &  159.5 &  \textbf{162.6 ↑ 3.12} \\
LLaMA-65B    &    3.733 &  \underline{4.673 ↑ 1.10} &    7.339 &  7.986 ↑ .746 &  3.363 &  \underline{3.756 ↑ .453} &  8.391 &  \underline{8.376 ↓ .018} &  175.2 &  172.4 ↓ 3.17 &  162.2 &  \underline{167.9 ↑  5.68} \\
Llama-2-70B  &    3.461 &  \textbf{3.995 ↑ .622} &    6.754 &  \textbf{7.372 ↑ .711} &  3.534 &  \textbf{3.770 ↑ .272} &  8.291 &  \textbf{8.242 ↓ .056} &  193.9 &  189.2 ↓ 5.36 &  175.9 &  178.2 ↑ 2.31 \\
Yi-34B       &    6.204 &  6.624 ↑ .490 &    7.321 &  \underline{7.713 ↑ .452} &  4.030 &  4.159 ↑ .149 &  8.547 &  8.475 ↓ .083 &  185.1 &  183.1 ↓ 2.34 &  180.7 &  176.1 ↓ 4.61 \\
\midrule
FLAC         &    95.09 &  95.08 ↓ .023 &    95.14 &  95.05 ↓ .094 &  95.06 &  94.95 ↓ .118 &  96.14 &  96.00 ↓ .161 &  81.20 &  80.96 ↓ .274 &  76.26 &  69.98 ↓ 7.24 \\
PNG          &    37.85 &  37.75 ↓ .115 &    36.71 &  37.40 ↑ .800 &  17.90 &  16.42 ↓ 1.71 &  33.15 &  33.08 ↓ .078 &  64.91 &  64.70 ↓ .245 &  86.30 &  89.47 ↑ 3.65 \\
Gzip         &    37.76 &  37.66 ↓ .115 &    36.59 &  37.29 ↑ .801 &  17.71 &  16.21 ↓ 1.73 &  33.07 &  33.00 ↓ .073 &  64.88 &  64.67 ↓ .243 &  86.29 &  89.46 ↑ 3.66 \\
\bottomrule
\end{tabular}

}
    \caption{Compression rate (\%, lower is better) on six testing data. \(\mathrm{Avg.} = Rate_{17-23} \) and \(2023 = Rate_{23}\). The performance difference (\(Rate_{23} - Rate_{17-22}\)) is indicated by an arrow followed by a number. ↑ indicates a higher (worse) rate on 2023. Assuming the rate changes approximately linearly, we could obtain an estimated future performance \(Rate_{24}\) by adding \(Rate_{23}\) to the rate difference observed in 2023. Models are highlighted in bold (1st best) and underlined (2nd best) based on their estimated future generalization.
    Models under and over 7B are compared separately.
    }
    \label{tab:big-tab}
    \vspace{-0.8em}
\end{table*}

\subsection{Generalization and Training Cutoff}

In this section, we focus on the LLaMA and Llama-2 series of models and analyze the correlation between their performance and cutoff date. Model performance is tested using wikitext and BBC news and the results are illustrated in the temporal space as shown in Figure \ref{fig:llama-time}.

In wikitext, there is a clear divergence in model performance that aligns closely with the cutoff dates. From 2017 to 2020 (the training period of both), LLaMA and Llama-2 exhibit comparable and steady performance. However, the divergence occurs after LLaMA's cutoff. LLaMA's performance begins to worsen rapidly after 2020, as shown by the sharp increasing trend of its compression rates. In contrast, Llama-2, thanks to its more up-to-date cutoff, exhibits a better generalization compared to LLaMA after 2020 and maintains the gradual performance decline even after its September 2022 cutoff. This performance-cutoff correlation is more obvious on the BBC News dataset. Both models achieve similar performance from Oct 2017 to Jan 2020 (the training period of both), but after the LLaMA cutoff, there is a significant increase in the compression rate, indicating a rapid fall-off in LLaMA's generalization on new and unseen data. On the contrary, Llama-2's performance remains consistent until its own cutoff in September 2022, where Llama-2 models also demonstrate a sharp decline after their corresponding cutoff.

These patterns suggest that whilst models may demonstrate similar performance on data from their training period, their ability to generalize on future data can differ significantly. This observation confirms data contamination issues in existing language model evaluations. Many benchmarks use data that overlaps with models' training periods, therefore potentially failing to accurately measure model generalization capabilities. As a result, our approach focuses on the performance on post training-period data, and includes the robustness dimension measured by the performance gap between the training and testing periods.

\begin{figure*}[t]
    \centering
    \includegraphics[width=\textwidth]{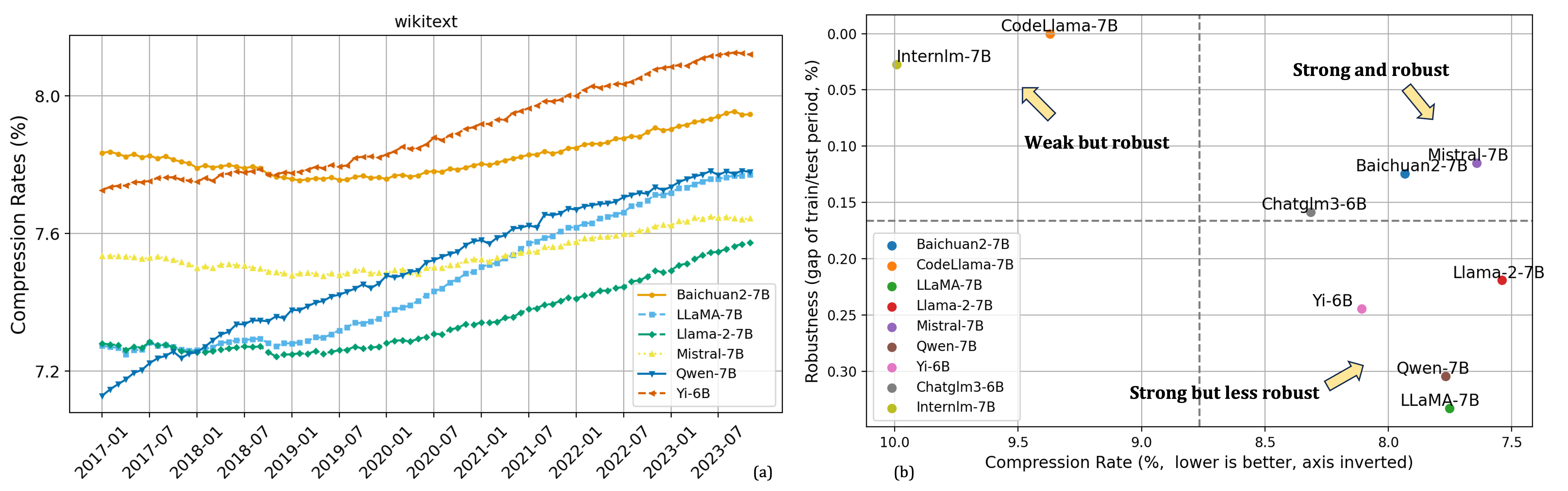}
    \vspace{-1em}
    \caption{(a) Compression rate on wikitext; (b) Robustness (gap between \(\mathrm{Rate_{23}}\) and \(\mathrm{Rate_{17-22}}\)) and performance (\(\mathrm{Rate_{23}}\)), tested on Wikitext. InternLM and CodeLlama are excluded from (a) for the sake of figure readability.}
    \vspace{-0.5em}
    \label{fig:model-comparison}
\end{figure*}

\subsection{Model Comparison}

In this section, we test and compare a wide selection of large language models on their generalization and robustness. In our experiments, we split the the test data into the training period (2017-2022) and the testing period (2023). 
Although many models do not report their exact cutoff dates, we choose 2023 as the testing period according to their release time (cf. Table \ref{tab:models}). As 2023 is rather recent, we believe it is not largely included in the pre-training stage. The results on our six test datasets are reported in Table \ref{tab:big-tab}. Firstly, we show their averaged performance over the entire period (i.e, \(\mathrm{Avg.}=Rate_{17-23}\)). Then, performance on the 2023 split is shown specifically (i.e., \(2023=Rate_{23}\)). Finally, the performance difference between the training and testing periods is reported as the measure for robustness (i.e., \(Rate_{23} - Rate_{17-22}\)). We use an arrow to represent the direction of performance change. An up arrow denotes that a model achieves a worse compression rate on 2023 (testing period) data compared to the period of 2017-2022 (training period), and vice versa.
To visually compare generalization and robustness, we place each model on the 2D performance-robustness space in Figure \ref{fig:model-comparison} (b). 
The x-axis represents performance on the 2023 testing period, with lower rates indicating stronger generalization on new data. The y-axis denotes the gap between training and testing performance, with smaller gaps denoting stronger robustness.
The relative position of each model reveals its capabilities. Models in the top right exhibit both strong generalization (low \(Rate_{23}\)) and high robustness (small training-testing gap). Models in the lower right achieve good generalization but less robustness over time. Models in the top left demonstrate weaker generalization but maintain robustness between periods.
We also illustrate compression rates over time in Figure \ref{fig:model-comparison} (a). Note that Qwen is not tested on multi-modal datasets as it does not reserve byte tokens in its vocabulary.

Our results reveal the following key findings. Firstly, models can demonstrate widely divergent performance on training versus testing data. For instance, LLaMA-65B's compression rate worsens 20\% on 2023 Wikipedia data compared to 2017-2022.
Secondly, CodeLlama excels on code data thanks to its additional training on code corpora but exhibits lower robustness compared to base Llama-2. This suggests that although further training on domain knowledge can lead to better domain capability, it may result in weaker generalization compared to the base model.
Thirdly, analyzing temporal trends provides insights on future generalization. In Figure \ref{fig:model-comparison} (a), Mistral and Baichuan2 display more gradual upward curves, suggesting better potential for future performance than models with steeper uptrends.
Finally, the 2-D visualization in Figure \ref{fig:model-comparison} (b) enables comparison of performance-robustness trade-offs. Overall, Mistral-7B achieves the most favorable balance among models under 7B and Llama-2-70B is the best among models over 7B. We include more figures of compression rate visualization in the temporal space and performance-robustness 2-D relations in Appendix \ref{more-comparison}.

\subsection{Comparison to Established Benchmarks}

To further analyze our compression-based evaluation method, we compare compression rates on our code and arXiv datasets versus scores on the established benchmarks HumanEval \cite{chen2021evaluating} and MMLU \cite{hendrycks2020measuring}. HumanEval is a benchmark focusing on programming tasks, and MMLU is a comprehensive evaluation consisting of academic tests. As shown in Table \ref{tab:compare-to-benchmarks}, compression performance correlates closely with models' ranking on these benchmarks. 
For academic tests, we compare our arXiv results to MMLU. Although arXiv contains primarily new scientific papers rather than traditional academic knowledge (like MMLU), it is the most relevant source among our collected datasets. 
Overall, this close correlation suggests our compression-based metric can be an effective method for model evaluation.

\begin{table}[t]
    \centering
    \resizebox{\columnwidth}{!}{
    \begin{tabular}{lcc|cc}
    \toprule
        Model & HumanEval & Code (ours) & MMLU & ArXiv (ours) \\
        \midrule
        CodeLlama & 33.5 (\#1) & 3.479 (\#1) & 36.9 (\#6) & 9.889 (\#6) \\
        LLaMA-7B & 11.6 (\#6) & 4.448 (\#5) & 35.1 (\#7) & 9.186 (\#5) \\
        Llama-2-7B & 12.8 (\#5) & 4.455 (\#6) & 45.3 (\#5) & 9.088 (\#4) \\
        Mistral-7B & 30.5 (\#3) & 4.043 (\#3) & 60.1 (\#1) & 8.372 (\#1)\\
        Qwen-7B & 32.3 (\#2) & 3.838 (\#2) & 58.2 (\#2) & 8.911 (\#2) \\
        Baichuan2-7B & 18.3 (\#4) & 4.293 (\#4) & 54.2 (\#3) & 9.018 (\#3) \\
        InternLM-7B & 10.4 (\#7) & 4.924 (\#7) & 51.0 (\#4) & 10.88 (\#7)\\
        \bottomrule
    \end{tabular}}
    \vspace{-0.5em}
    \caption{Relation between compression rate (\(Rate_{23}\), \%) and benchmark score. pass@1 for HumanEval, 5-shot accuracy for MMLU. Results for HumanEval and MMLU are  from the official paper or website of each respective model.}
    \label{tab:compare-to-benchmarks}
    \vspace{-1em}
\end{table}

\subsection{Generalization on Different Data Sources}

Our findings also show insights into how large language models generalize across diverse datasets. Firstly, models face more challenges with 2023 Wikipedia, news, and code data but not with arXiv or multi-modal data. Gzip compression rates are lower for Wikipedia and code in 2023, indicating more repetitive patterns. However, language models perform worse on 2023 data, revealing difficulties in generalizing to new time-sensitive textual data. Surprisingly, for arXiv, models maintain or even improve performance on 2023 data, perhaps due to consistent writing styles in academic papers. Secondly, large language models struggle with pure byte streams, as seen in their compression rates on multi-modal datasets, suggesting difficulty in identifying byte patterns. Future work will include multi-modal models in byte stream compression. Finally, on time-sensitive textual datasets, models exhibit distinct trends. In Figure \ref{fig:llama-time}, wikitext performance gradually declines post-cutoff, suggesting slow changes in encyclopedic knowledge for model generalization. In contrast, news dataset performance sharply drops, indicating the rapid content changes that are prevalent in news. We include more figures demonstrating how models generalize on the news, code, image, and audio datasets in Appendix \ref{more-trend-over-time}.

\begin{table*}[t]
    \centering
    \resizebox{\textwidth}{!}{
    \begin{tabular}{lcccccccccccc}
\toprule
& \multicolumn{3}{c}{2K} & \multicolumn{3}{c}{4K} & \multicolumn{3}{c}{8K} & \multicolumn{3}{c}{2K+SW} \\
\cmidrule(lr){2-4}
\cmidrule(lr){5-7}
\cmidrule(lr){8-10}
\cmidrule(lr){11-13}
Model & Rate & Time(s) & Mem(mb) & Rate & Time(s) & Mem(mb) & Rate & Time(s) & Mem(mb) & Rate & Time(s) & Mem(mb) \\
\midrule
Chatglm3-6B & 8.314 & 110 & 13563 & 8.144 & 96 & 15175 & 8.064 & 98 & 18397 & 7.965 & 422 & 13562 \\
Baichuan2-7B & 7.932 & 134 & 19548 & 7.831 & 146 & 24843 & - & - & - & 7.731 & 518 & 19550 \\
Qwen-7B & 7.768 & 100 & 20043 & 7.662 & 93 & 25347 & 7.606 & 92 & 35951 & 7.555 & 401 & 20041 \\
Mistral-7B & 7.642 & 118 & 15487 & 7.509 & 108 & 16639 & 7.462 & 107 & 18945 & 7.385 & 475 & 15487 \\
Llama-2-7B & 7.539 & 114 & 15615 & 7.409 & 101 & 18303 & - & - & - & 7.283 & 421 & 15615 \\
\bottomrule
\end{tabular}
    }
    \vspace{-0.5em}
    \caption{Compression rate (\%) comparison for context sizes on the 2023 split of wikitext.}
    \vspace{-0.5em}
    \label{tab:context}
\end{table*}

\subsection{Context Size and Performance}
\label{context-size}

Context size is a key factor in compression algorithms. Typically, a larger context allows algorithms to identify useful patterns in a broader scope, leading to better performance. This principle also applies to large language models, where an extensive context is crucial for complex reasoning, long-dependency question-answering, and coding tasks \cite{touvron2023llama2}. However, larger contexts inevitably require more computing and memory, so a trade-off has to be made for compression algorithms and large language models. In this section, we examine how the context size affects the compression rate of models and their computational costs.

In Table \ref{tab:context}, we present the performance of models across different context sizes: 2K, 4K, 8K, and 2K with a sliding window (2K+SW). For the first three configurations, compression is performed chunk by chunk, with no overlap between chunks. In the sliding window setup, we use a step size of 512, where only the probabilities of the last 512 tokens are calculated in each step, and the first 1536 tokens serve to provide context. This method requires approximately four times the computing resources compared to a 2K context but is expected to yield better performance. Our key findings are as follows. Firstly, there is a general trend of improved compression rates as the context size increases from 2K to 4K and then to 8K across all models. However, the extent of improvement diminishes with larger contexts, with the performance gain from 2K to 4K being generally more significant than that from 4K to 8K.
Secondly, there is a noticeable increase in memory usage as the context size grows. Mistral and ChatGLM3 show impressive memory efficiency in dealing with large context, perhaps thanks to their unique attention implementation \cite{jiang2023mistral}. In contrast, large contexts do not consume more time to finish compressing. This is because there are fewer chunks when we work with larger contexts and a chunk-by-chunk approach. Lastly, the 2K context size with a sliding window (2K+SW) consistently outperforms the larger static context sizes in all models. This might result from the continuous context provided by the sliding window, which we believe is crucial for the initial part of each chunk.

\begin{table}[t]
    \centering
    \resizebox{\columnwidth}{!}{
    \begin{tabular}{lcccc}
    \toprule
    Model & Vocab\_size &  \#Tokens &  BPT &  BPB \\
    \midrule
    Qwen-7B           &          152K & 12382K &     2.7511 &     0.6215 \\
    Baichuan2-7B &          126K & 12824K &     2.7135 &     0.6346 \\
    Chatglm3-6B  &          65K & 13531K &     2.6951 &     0.6652 \\
    Llama-2-7B     &          32K & 14324K &     2.3086 &     0.6032 \\
    Mistral-7B        &          32K & 14006K &     2.3929 &     0.6113 \\
    \bottomrule
    \end{tabular}}
    \vspace{-0.5em}
    \caption{Comparison of models with different tokenization implementations on the 2023 split of wikitext.}
    \vspace{-0.8em}
    \label{tab:tokenizer}
\end{table}

\subsection{Tokenization and Performance}

Tokenization aims to represent linguistic input more efficiently before language modeling, and thus is often considered as a form of pre-compression \cite{deletang2023language}. In this section, we analyze how the implementation of tokenization affects models' compression rates. We employ two metrics in Table \ref{tab:tokenizer}: bits per token (BPT) and bits per character (BPC), indicating the number of bits required to represent a token or character, respectively. A lower BPT or BPC suggests more efficient data compression. However, they focus on different aspects: BPT assesses the model's efficiency at the token level, whilst BPC evaluates performance at the character level. Additionally, the table includes the total number of tokens, reflecting the resulting token count for the entire test dataset.

Our key findings are as follows: Firstly, models optimized for multilingual scenarios often have a significantly larger vocabulary size than monolingual models. For example, Qwen, as a multilingual model, has a 5 times larger vocabulary than that of Llama-2. Secondly, a large vocabulary generally leads to fewer tokens, which means tokenizers perform better on pre-compression. Thirdly, a large vocabulary may lead to increased complexity in token-level predictions. As shown in Table \ref{tab:tokenizer}, models with larger vocabularies tend to achieve a higher BPT in general, indicating the challenge of token prediction with more fine-grained tokenization.
Note that this analysis was conducted on purely English data, which inherently favors English models, especially in tokenization analysis. Experiments with multilingual test data should be included in future works.

\section{Conclusion}

We proposed to evaluate large language models through lossless data compression across time on generalization and robustness. Our method measures compression rates on new, unseen data that is created after a model's training cutoff, as a metric for generalization. It also quantifies the gap between the training and testing periods as a metric of robustness. The proposed method avoids data contamination and the potential interference of different prompts in existing benchmark-based evaluation. We provided an extensive analysis for 14 representative large language models, and also investigated the impact of context size and tokenization implementations.


\section{Impact}

This paper focuses on evaluation methods for large language models. 
There are many potential societal consequences of our work, none which we feel must be specifically highlighted here.

\nocite{li2023LatestEval}
\nocite{li2023compressing}
\nocite{li2023estimating}
\nocite{li2023unlocking}

\bibliography{icml2024}
\bibliographystyle{icml2024}

\newpage
\appendix
\onecolumn
\section{Data License}
\label{license}

Wikipedia and arXiv are under the CC BY-SA license which grants free use of their content. BBC data including news, images, and radio are under the Educational Recording Agency (ERA) license, which allows educational usage of their content. For the code datasets, we only collect data from GitHub projects under MIT or BSD licenses. For the podcast data in the audio dataset, we collect podcast audio from YouTube that are under the Creative Commons license, which allows free use of their content.

\section{Test Data Analysis}
\label{how different}

\begin{figure*}[h]
    \centering
\includegraphics[width=0.8\textwidth]{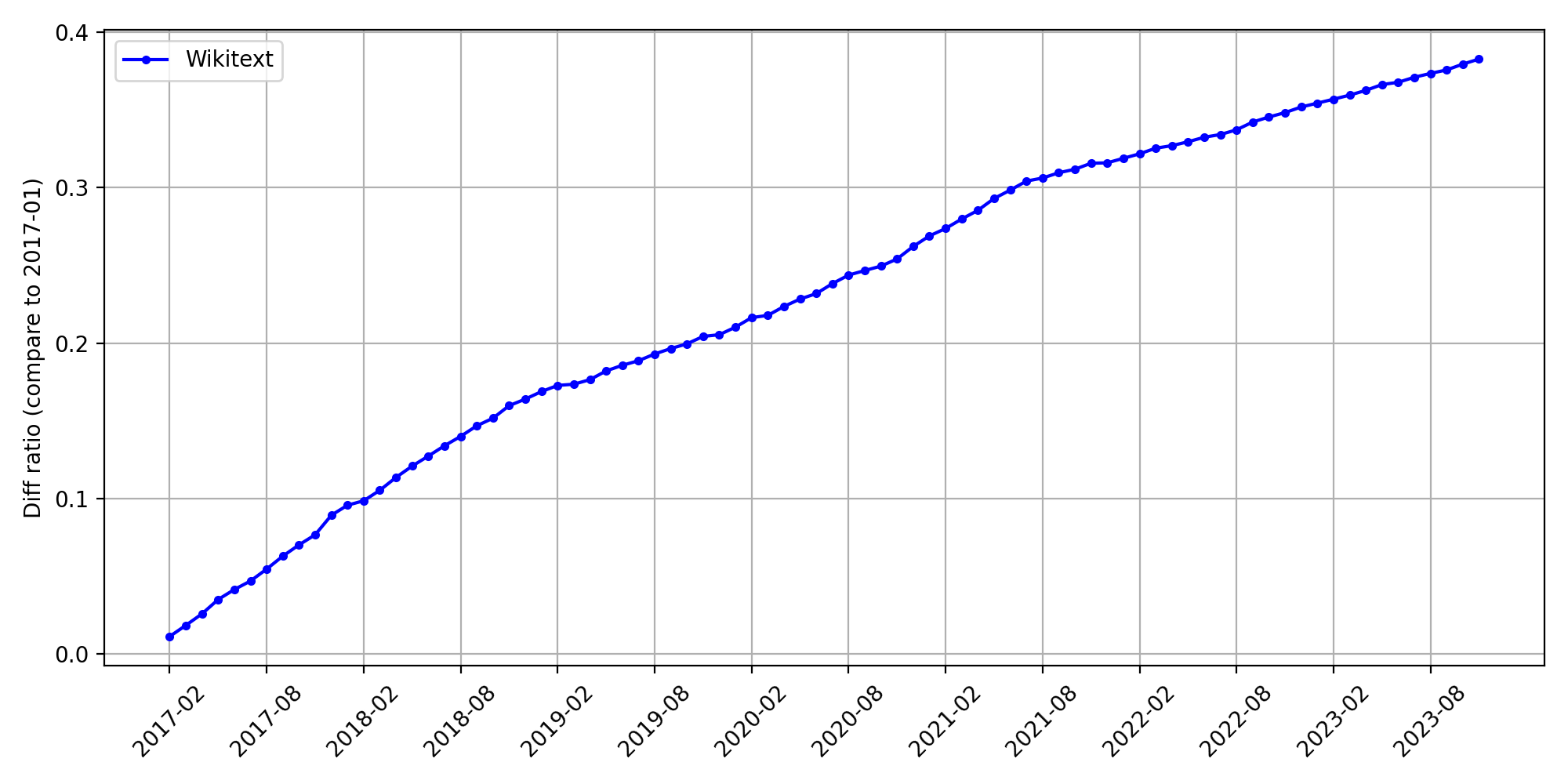}
    \caption{The differences in monthly Wikipedia data.}
    \label{fig:wiki_diff}
\end{figure*}

Here we investigate how different our test data is when collected at monthly intervals. We use the Gestalt pattern matching algorithm (implemented by Python \texttt{difflib.SequenceMatcher}) to compute text similarity. For Wikipedia, we perform the comparison article-by-article, while for other sources such as news, code, and arXiv, we concatenate all data together before computing the similarity. All sources demonstrate large differences over 0.9 (similarity below 0.1) except Wikipedia. The average differences are 0.98, 0.91, 0.96 for BBC news, Code, and arXiv respectively. As shown in Figure \ref{fig:wiki_diff}, since we are monitoring a specific set of Wikipedia articles, these wiki pages change approximately linearly over time.

\section{Generalization on Different Domains}
\label{more-trend-over-time}

\begin{figure*}[t]
    \centering
    \includegraphics[width=\textwidth]{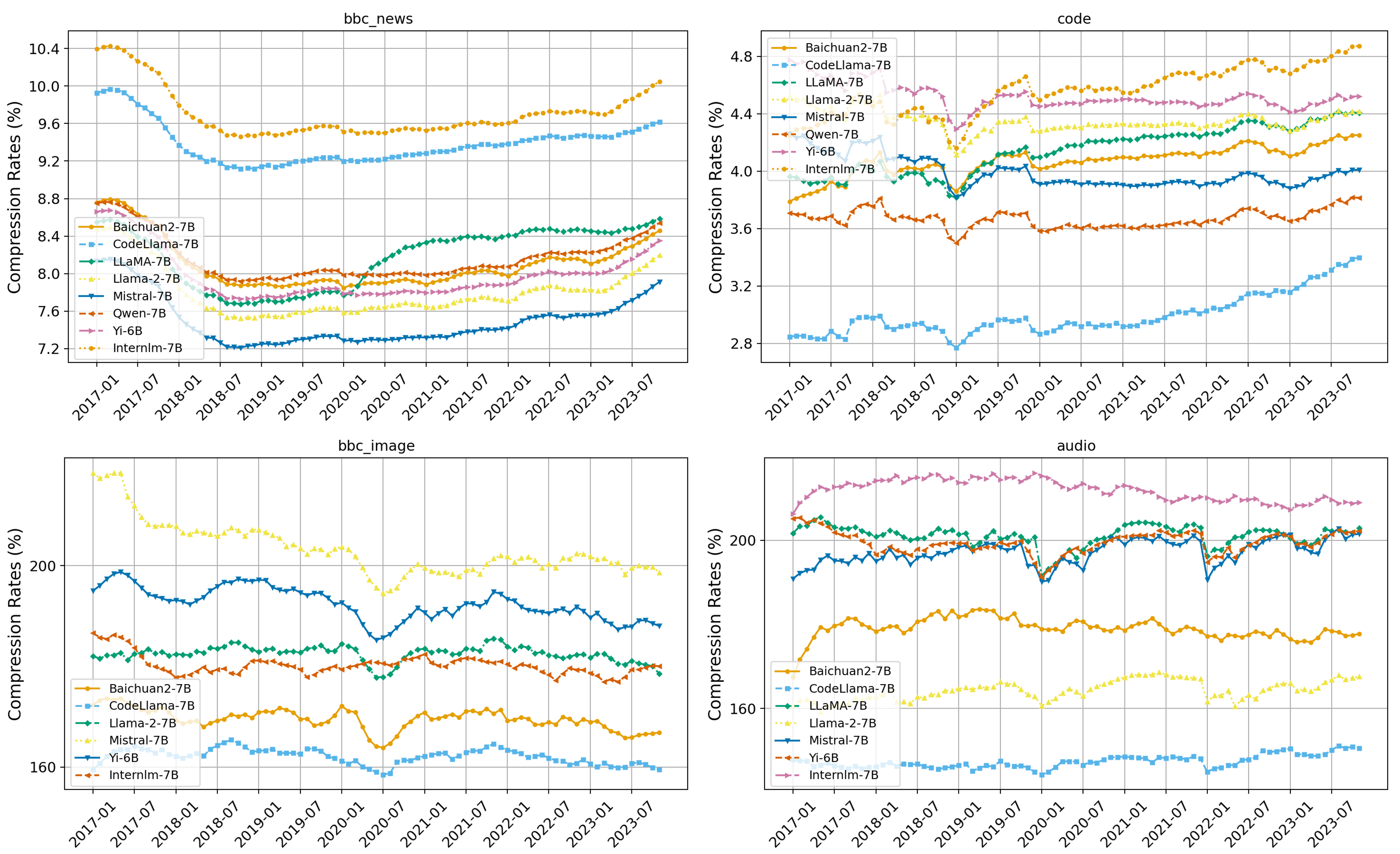}
    \caption{Compression rates change over time on different domains.}
    \label{fig:more-trend}
\end{figure*}

We mentioned that large language models behave quite differently across domains. Here, we add more visualizations showing how well models generalize over time on different test datasets. As shown in Figure \ref{fig:more-trend}, we compare all models under 7B parameters, excluding larger models to avoid a cluttered figure. The key findings we could derive from the temporal visualizations are the trends and slopes of model performance in the recent period. For example, on the BBC news dataset, all models demonstrate an increasing trend on the 2023 split. Notably, their increasing slopes seem similar overall. This suggests the models have comparable robustness, perhaps because news topics often change randomly and dramatically. In contrast, on the GitHub code dataset, the increasing slopes of CodeLlama and InternLM are more significant than other models after 2023, highlighting their potential weaker robustness on coding tasks.

\section{Model Comparison on Different Sources}
\label{more-comparison}

\begin{figure*}
    \centering
    \includegraphics[width=\textwidth]{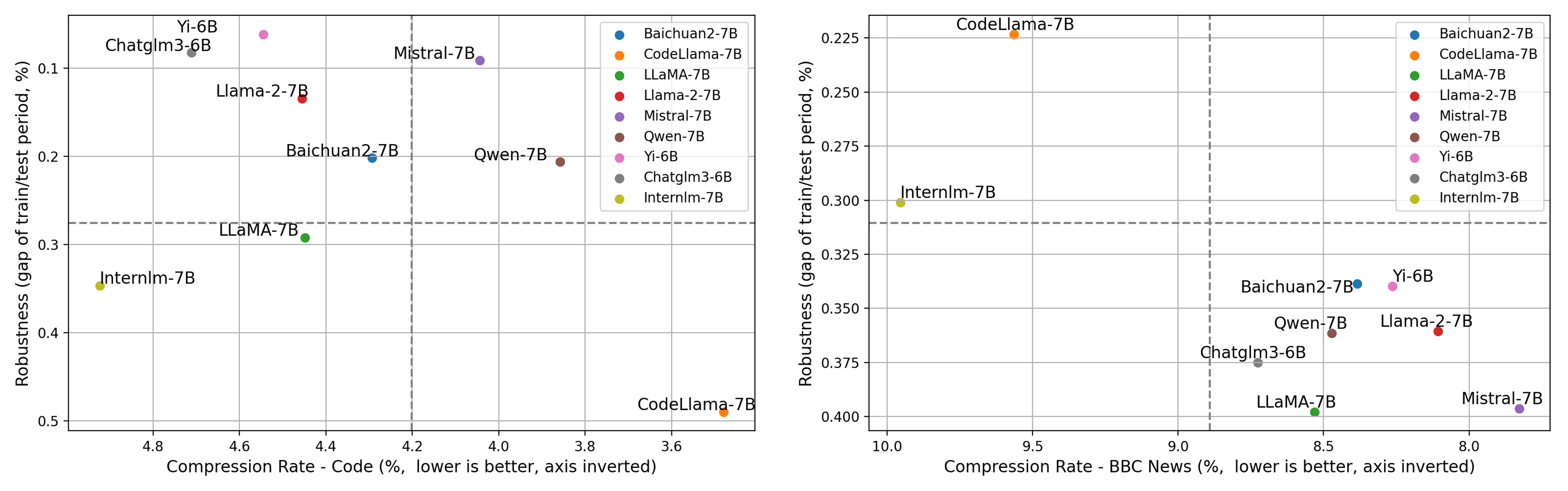}
    \caption{The relation between performance and robustness on GitHub code (left) and BBC news (right).}
    \label{fig:more-comparison}
\end{figure*}

We add the performance and robustness 2D visualizations of GitHub code and BBC news datasets, as shown in Figure \ref{fig:more-comparison}.

\end{document}